\documentclass[twoside]{article}

\pdfoutput=1
\usepackage{xspace}
\usepackage{url}
\usepackage{microtype}
\usepackage{graphicx}
\usepackage{subcaption}
\usepackage{booktabs} 
\usepackage{enumitem}
\usepackage{dsfont}
\usepackage{multirow}
\usepackage{wrapfig}

\usepackage{amsmath}
\usepackage{amssymb}
\usepackage{mathtools}
\usepackage{amsthm}

\usepackage[accepted]{aistats2025}
\usepackage[round]{natbib}

\begin{document}

\newcommand{\framework}{M\textsuperscript{2}AD\xspace}
\newtheorem{proposition}{Proposition}
\newcommand\he[1]{{\textbf{[ZH: #1]}}} 
%
\runningtitle{\framework: Multi-Sensor Multi-System Anomaly Detection}

%
\runningauthor{Alnegheimish, He, Reimherr, Chandrayan, Pradhan, D'Angelo}

\renewcommand{\thefootnote}{\fnsymbol{footnote}}

\twocolumn[
\aistatstitle{\framework: Multi-Sensor Multi-System Anomaly Detection through Global Scoring and Calibrated Thresholding}
\aistatsauthor{Sarah Alnegheimish\footnotemark[1] \And Zelin He\footnotemark[1] \And  Matthew Reimherr\footnotemark[2]}
\aistatsaddress{MIT \And Pennsylvania State University \And Pennsylvania State University \\ Amazon} \vspace{-11pt}
\aistatsauthor{Akash Chandrayan \And Abhinav Pradhan \And Luca D'Angelo}
\aistatsaddress{Amazon \And Amazon \And Amazon } ]

\begin{abstract}
    With the widespread availability of sensor data across industrial and operational systems, we frequently encounter heterogeneous time series from multiple systems. Anomaly detection is crucial for such systems to facilitate predictive maintenance. However, most existing anomaly detection methods are designed for either univariate or single-system multivariate data, making them insufficient for these complex scenarios. To address this, we introduce \framework, a framework for unsupervised anomaly detection in \textit{multivariate} time series data from \textit{multiple} systems. \framework employs deep models to capture expected behavior under normal conditions, using the residuals as indicators of potential anomalies. These residuals are then aggregated into a global anomaly score through a Gaussian Mixture Model and Gamma calibration. We theoretically demonstrate that this framework can effectively address heterogeneity and dependencies across sensors and systems. Empirically, \framework outperforms existing methods in extensive evaluations by 21\% on average, and its effectiveness is demonstrated on a large-scale real-world case study on 130 assets in Amazon Fulfillment Centers. Our code and results are available at \url{https://github.com/sarahmish/M2AD}.
\end{abstract}

\section{Introduction}
\label{sec:introduction}
Over the past several years, there has been a tremendous shift towards sensor-based monitoring of mechanical and electrical assets, whether in manufacturing, healthcare, logistics, or energy management~\citep{healthcare2022abdulmalek, windenergy2020liu, tahan2017gasturbine, manufacturing2019cao}. These sensors generate a significant amount of data in real time, reinforcing the need for automated solutions for anomaly detection to reduce manual labor. Moreover, it emphasizes the importance of deploying models that are capable of processing large amounts of data concurrently and with high accuracy.

As the number of assets (machines) increases and data sources become more diverse, sensor-based monitoring provides richer, more complex information. For one, as the variety of sensors increases, we now have a multitude of metrics coming from various sources to assess the health condition of assets.
For example, a conveyor belt in a warehouse is monitored through multiple sensors that measure temperature and vibrations at various locations of the belt, and the electric current needed by the motors to drive the conveyor.
Moreover, additional measurements provide valuable insights into an asset's throughput and performance. For instance, in e-commerce settings such as Amazon's fulfillment centers, we can track how many items a machine processes. This information is particularly useful in domains where activity levels are highly seasonal and do not follow simple patterns. By integrating data from various systems, we gain a more comprehensive view of asset utilization and health, enabling more informed decisions about performance and maintenance.

\footnotetext[1]{Work done during Amazon internship.}
\footnotetext[2]{Corresponding author mreimherr@psu.edu.}

\renewcommand{\thefootnote}{\arabic{footnote}}

Monitoring the condition of assets is crucial for ensuring reliable operation and minimizing unplanned downtime. In this context, \textit{unsupervised} time series anomaly detection has emerged as a key approach to facilitate condition-based monitoring, allowing for the identification of anomalies without requiring previously labeled data. For example, \citet{usad2020audibert} propose adversely trained autoencoders to generate anomaly scores, however, the authors apply manual thresholding which is both impractical and not scalable for real-world applications. On the other hand, \citet{lstmdt2018hundman} address this issue by developing a non-parametric thresholding model, yet it is designed for detecting anomalies in univariate time series, which could significantly increase the number of false alarms in multivariate data. 


Detecting anomalies in multivariate time series presents additional challenges in real-world industrial settings. To be effective in practice, the model needs to meet key operational requirements. First, it must raise alarms judiciously to optimize technician workload and build trust in the system. Second, for diagnostic purposes, it is crucial to provide interpretability, allowing operators to identify which sensor in which system triggered the alarm. This gives the respective technician context and helps reduce the investigation time.

To address these challenges, we propose a multi-sensor multi-system anomaly detection framework, \framework, for time series data. The framework uses a prediction model to learn the expected pattern of the signal, we then use the discrepancies between the predicted and actual signals to compute a global anomaly score that factors in all sensors from available systems. The framework learns the appropriate threshold value estimated from anomaly scores seen during the training phase. Moreover, to preserve the element of interpretability, the system produces an ordered list of most contributing sensors to the anomaly score.
We summarize our contributions below:
\begin{itemize}[topsep=0pt, itemsep=0pt, leftmargin=*]
    \item Introduce \textbf{\framework}, an unsupervised \textbf{multi-sensor multi-system} time series \textbf{anomaly detection} method. \framework leverages an LSTM model to learn patterns from sensor observations and covariates, identifying discrepancies between expected and observed behavior. These sensor-wise discrepancies are aggregated into a global anomaly score through a novel combination of a Gaussian Mixture Model and Gamma calibration, effectively addressing heterogeneity and dependencies across sensors and systems.
    \item Provide \textbf{theoretical support} for \framework using learning theory, demonstrating that it provably reduces false detection and improves error distribution calibration, particularly in scenarios with dependencies and heterogeneity across sensors within multiple systems.
    \item Conduct a comprehensive \textbf{empirical evaluation} of \framework on three public multivariate time series datasets, benchmarking it against eight state-of-the-art deep learning models. Our model achieves a 21\% improvement over existing methods. Additionally, we validate \framework through a real-world case study using proprietary data of 130 assets from an Amazon fulfillment center, evaluating its effectiveness.
\end{itemize}
\section{Related Work}
\label{sec:related_work}
There has been a continuous effort to develop new unsupervised time-series anomaly detection methods using the latest models and architectures~\citep{zamanzadeh2024survey1, choi2021survey2, braei2020survey3, kim2024contrastive}. Relevant work can broadly be categorized into the following:

\noindent\textbf{Univariate Anomaly Detection Methods.}
There is a variety of prediction-- and reconstruction-based models for unsupervised anomaly detection in univariate time series~\citep{wong2022aer, ren2019azure, munir2019deepant}. ARIMA~\citep{box1970arima} is a classic auto-regressive model to find the residuals between the real and predicted time series. Similarly, \citet{lstmdt2018hundman} utilizes an LSTM with a non-parametric threshold to locate the position of anomalies. On the other hand, \citet{geiger2020tadgan} proposes a reconstruction-based model using GANs to find anomalies in time series through the reconstruction error and discriminator scores.
Univariate models neglect the correlation between various channels and cannot be scaled efficiently, making them unsuitable for multivariate time series. 

\noindent\textbf{Multivariate Anomaly Detection Methods.}
Reconstruction-based methods are also utilized in multivariate time series anomaly detection~\citep{xu2022anomalytransformer, tuli2022tranad, wu2023timesnet, xu2024fits, nam2024dualtf, fang2024tfmae}. 
\citet{malhotra2016lstmautoencoder} develops an LSTM autoencoder and calculates an anomaly score using reconstruction error. \citet{usad2020audibert} applies an autoencoder model with a two-phase adversarial training scheme.
\citet{li2021interfusion} uses a hierarchal variational autoencoder with explicit low-dimensional inter-metric and temporal embeddings.
OmniAnomaly uses RNNs to learn the normal representation of the data,
utilizing reconstruction probabilities and an automatic threshold~\citep{smd2019su}.
Except for the last model, all the proposed solutions construct a score for anomalies without properly configuring the appropriate threshold, requiring ad-hoc tuning of threshold values. In contrast to \citet{smd2019su}, we provide a simpler prediction-based model to generate a multivariate time series and derive a data-driven threshold.

\noindent\textbf{Anomaly Detection on Multiple Systems.}
Working with multiple data sources has become critical as the number of monitoring devices increase.
\citet{cnnrnn2019canizo} extracts features for each sensor independently using CNNs. The features are then concatenated and passed to an RNN to classify if they are anomalous or not. Extracting features of each sensor independently ignores the relationship between different sensors. 
Moreover, \citet{park2018vae} proposes a reconstruction-based approach using variational autoencoders for multiple sensory data from different systems using LSTM layers for robotics. The algorithm computes the anomaly score from a global negative log-likelihood of an observation, making it is difficult to interpret which sensor triggered the anomaly.


\section{\framework: Multi-Sensor Multi-System Anomaly Detection}
\label{sec:systen}

\begin{figure}
    \centering
    \includegraphics[width=1\linewidth]{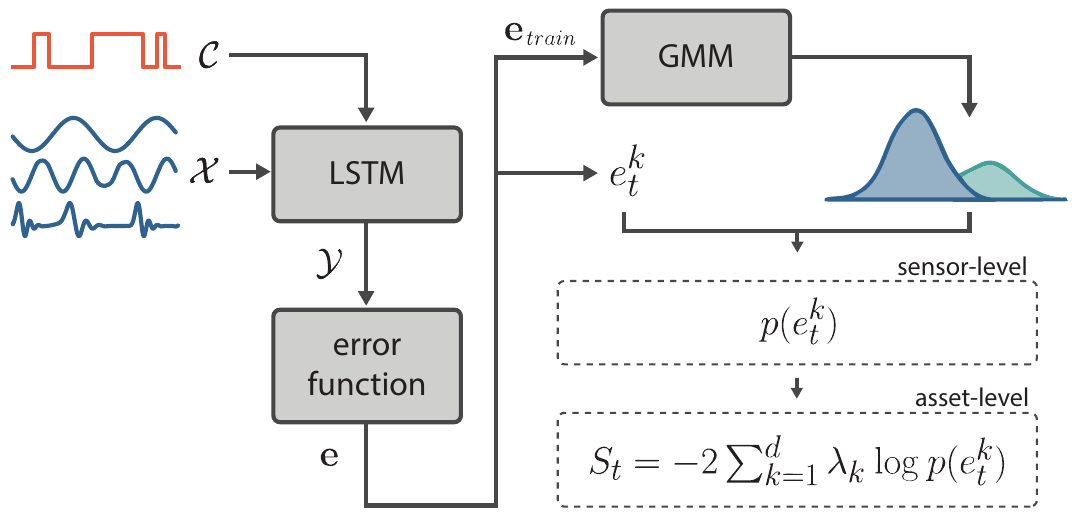}
    \caption{Overview of \framework workflow.
    The frameworks trains an LSTM to predict $\mathcal{Y}$, which models the expected pattern of $\mathcal{X}$. Then, \framework calculates the errors ($\mathbf{e}$) between the observed and expected signal for each sensor. Lastly, we use a Gaussian Mixture Model to find p-values and construct a global anomaly score for a particular asset from all its sensor information.}
    \label{fig:framework}
\end{figure}

Given an asset, our objective is to identify if any anomalous behaviour is observed.  Figure~\ref{fig:framework} depicts an overview of our Multi-system Multivariate Time Series Anomaly Detection (\framework) framework.

\subsection{Problem Statement}
Consider anomaly detection for an asset, where a set of sensors observes multivariate time series data from potentially different systems. For simplicity of notation, we denote the combined observations as a single multivariate time series \(\mathcal{X} = \{x_1, \dots, x_T\}\), where \(x_t \in \mathbb{R}^d\) and \(d\) represents the total number of sensors. Optionally, we have covariates \(\mathcal{C} = \{c_1, \dots, c_T\}\) which is a categorical time series representing the status of the asset.
Our goal is to identify anomalous regions by finding $\mathcal{A} = \{t: z_t = 1\}$ in a given asset where $z_t \in \{0, 1\}$. Note that we do not have any prior information on the existence of anomalies, in fact $\mathcal{A}$ is often an empty set.

\framework, is composed from two phases: training and inference. We denote $\mathcal{D}_{train} = \{({x}_t, c_t)_{i=1}^n\}$  with $n < T$ as the training dataset, where we assume $\mathcal{D}_{train}$ is free from anomalies and is operating normally.
The remainder of the dataset will be used for testing $\mathcal{D}_{test} = \{({x}_t, c_t)_{i=n}^T\}$.

\subsection{Time Series Prediction}
The first step of the model is to model what the behavior of the time series should look like under normal conditions.
To train the model, we use the training dataset $\mathcal{D}_{train}$.
We denote $\mathcal{Y}$ as the predicted time series, learned from the sensor-- and status-type time series. 
We then aim to find anomalies in our inference data $\mathcal{D}_{test}$ where it may or may not have anomalies.
Multiple models can be used to estimate $\mathcal{Y}$ from $\mathcal{X}$ and covariates $\mathcal{C}$~\citep{taylor2018prophet, das2024timesfm, rasul2023lagllama, ansari2024chronos}.
We use a Long Short-Term Memory (LSTM) model to capture the temporal pattern of the signal~\citep{hochreiter1997lstm}. Moreover, having a multivariate prediction helps us capture the dependencies across multiple variables:
\begin{equation*}
    {y}_{t+1} = f({x}_{w_t}, c_{w_t}),
\end{equation*}
where $w_t$ refers to the historical observations we want to include at time $t$ and window size $w$, namely for input $x_{w_t}$ we can express it as $x_{w_t} \coloneq x_{t\dots t-w}$. 
We use the squared $L_2$ norm to compute the loss with respect to the sensor-type time series $\|{x}_{t} - {y}_{t}\|_2^2$.

\subsection{Discrepancy Calculation}
\label{sec:error}

When calculating the discrepancy $\mathbf{e} = \{e_1, e_2, \dots, e_T\}$, we want a function that highlights the possible occurrence of an anomaly, this is highly dependent on the behavior of the time series and we deem as being sensor-specific. Here, we propose two methods for finding this gap. 
\citet{lstmdt2018hundman} proposes a simple point-wise error to capture the differences $e_t = |x_t - y_t|$. Note that by taking the absolute value, we are not interested in the direction of the error but only the magnitude.

\begin{figure}
    \centering
    \includegraphics[width=0.6\linewidth]{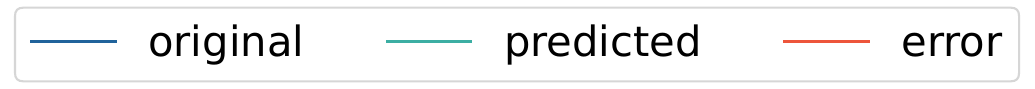}
    \includegraphics[width=1\linewidth]{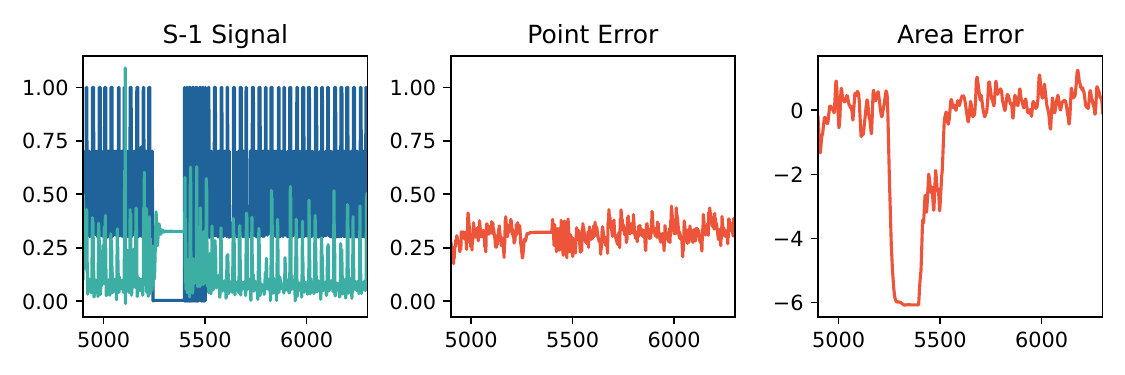}
    \caption{Example of S-1 signal from SMAP dataset. (left) depiction of the original time series and predicted one. (center) error signal using point-wise difference. (right) error signal using area difference.}
    \label{fig:example}
\end{figure}

In many applications, whether the expectation is higher/lower than what is actually observed matters. For example, in temperature data, the model should be sensitive to whether there is a substantial increase. However it should not consider drops in temperature as typically that is not problematic. 
Moreover, point-wise error is sensitive to singular extreme values, which makes them unsuitable to very noisy systems. To factor in these nuances, we use an area difference error function:
\begin{equation*}
    e_t = \frac{1}{2l} \int_{t-l}^{t+1} (x_t - y_t)\; dt,
\end{equation*}
where $l$ is the number of time points to include before and after the time $t$.
Figure~\ref{fig:error} shows an example of computing the difference between the original and predicted signals using point and area errors respectively. The location of the anomaly in the area error signal is more apparent than the point error signal.

Choosing between point and area error depends on the characteristics of the data and anomaly.
With point errors, one can effectively find out-of-range anomalies by measuring deviations in individual data points.
On the other hand, area error is better at discovering contextual anomalies by capturing deviations in patterns, even when individual values are within expected ranges.

\subsection{Anomaly Score}



Once we obtain the discrepancy measure for each sensor, it remains to design a global anomaly score for the whole asset. We frame the detection task as a multiple testing problem, where the null hypothesis $H_0$ assumes that none of the sensors exhibit anomalous behavior at time $t$. Formally, for an asset with $d$ sensors, the null hypothesis is:
\[
H_0:  e_t^k \sim f_{\text{normal}}^k \text{ for all } k = 1, \dots, d.
\]
Determining an anomaly score consists of two steps: first calculating sensor-level scores and then aggregating scores to form a global anomaly score for the asset.

\noindent\textbf{Sensor-Level Anomaly Score.}
For each sensor, we model the distribution of training error values using a Gaussian Mixture Model (GMM). Different sensors, especially those from different systems, often show distinct multimode error distributions due to various operational conditions. The GMM allows us to account for such a variation by fitting a multimode distribution. The GMM for sensor \(k\) is defined as follows:
\[
f^{k}(x) = \sum_{j=1}^{m_k} p_j \cdot \mathcal{N}(x \mid \mu_j, \sigma_j^2),
\]
where \(\mathcal{N}(x \mid \mu_j, \sigma_j^2)\) is the normal density with mean \(\mu_j\) and variance \(\sigma_j^2\), and \(p_j\) is the weight of the \(j\)-th component. Here, \(m_k\) represents the number of components in the mixture for sensor \(k\). See Appendix \ref{choice_of_mk} for more details on choosing $m_k$.

We estimate the parameters \(\mu_j\) and \(\sigma_j\) from the training errors, which represents normal operational behavior. To determine whether an observed error score \( e_t^k \) at time \(t\) is anomalous, we calculate a two-sided p-value using the cumulative distribution function (CDF) of the GMM. The CDF, based on the estimated parameters, is given by:
\[
F^{k}(x) = \sum_{j=1}^{m_k} p_j \cdot \Phi\left(\frac{x - \hat{\mu}_{j}}{\hat{\sigma}_{j}}\right),
\]
where \(\Phi\) is the standard normal CDF. The two-sided p-value is computed as $p(e_t^k) = 2 \min \{ F^{k}(e_t^k), 1 - F^{k}(e_t^k) \}$. A small p-value indicates an anomalous pattern is observed on sensor $k$. 
Note that with unidirectional errors, it is sufficient to conduct an upper-- or lower-tailed test.

\noindent\textbf{Asset-Level Anomaly Score.}
Once the sensor-level p-values have been obtained, we aggregate them to form a global anomaly score for the asset. One widely used aggregation method is the Fisher's method~\citep{fisher1928statistical}, which calculates the global anomaly score \( S_t \) for an asset with \(d\) sensors at time \(t\) is calculated as:
\[
S_t = -2 \sum_{k=1}^{d} \lambda_k \log p(e_t^k),
\]
where \( p(e_t^k) \) is the p-value from sensor \(k\) at time \(t\), and \(\lambda_k\)'s are sensor-specific or uniform weights. If all the p-values are independent, with uniform weights, $S_t$ follows a $\chi^2(2d)$ distribution under the null hypothesis. However, the p-values from different sensors may be heavily dependent on each other as this is a multi-sensor system monitoring the same asset. Therefore, we adjust classical Fisher’s method by fitting a Gamma distribution \(\Gamma(\alpha, \theta)\)  using p-values obtained from the training phase. Specifically, the shape and scale parameters \(\hat{\alpha}\) and \(\hat{\theta}\) are estimated from the training data by matching the moments of the aggregated log-transformed p-values:
\[
\hat{\alpha} = \frac{\left( E_{n}[S_t] \right)^2}{\text{Var}_{n}[S_t]}, \quad \hat{\theta} = \frac{\text{Var}_{n}[S_t]}{E_{n}[S_t]}.
\]
To classify a time point \(t\) as anomalous, we compare the global anomaly score \(S_t\) to a threshold derived from the fitted Gamma distribution. A time point is classified as anomalous if the global anomaly score exceeds a pre-defined threshold:
\[
z_t\coloneq \mathds{1}[S_t > \gamma], 
\]
where \(\gamma\) is the threshold value derived from the Gamma distribution \(\Gamma(\hat{\alpha}, \hat{\theta})\) for a desired significance. 
By following this two-step process, we can robustly combine sensor-level anomaly scores and detect anomalies at the asset level.

\subsection{Interpretability of Anomaly Score}
The modular design of \framework helps us verify which sensor from which system flagged an anomaly.
Having this level of interpretability is key to adoption at Amazon warehouses. This is because on-site technicians need to investigate and examine part of the asset that raised an alarm.
To achieve this, we simply find the most contributing sensor to the anomaly score at time $t$:
\[
    \arg \max_k\; \frac{\lambda_k \log p(e_t^k)}{S_t}.
\]
Top contribution can belong to different systems. In practice, we show the top 5 contributing sensors alongside their score and their respective systems.

\section{Properties of \framework Scoring}
In this section, we demonstrate the statistical properties of the anomaly score obtained from \framework.

\subsection{Improvement in Error Quantification}
One of the key feature of M\textsuperscript{2}AD in the sensor-level scores is the use of GMM model instead of the widely used univariate Gaussian distribution. If for one system the underlying true distribution of the anomaly score is well approximated by a mixed Gaussian, 
the univariate Gaussian approach will be heavily biased. Here we characterize the bias introduced due to such a misspecification. Consider the case where the true model distribution, $\mathbb{P}_*$, is a mixture of a two-component Gaussian location mixture given by
\begin{align}
\label{eq:true_model}
\mathbb{P}_*=\frac{1}{2} \mathcal{N}\left(\mu^*(1+\rho), \sigma^2\right)+\frac{1}{2} \mathcal{N}\left(\mu^*(1-\rho), \sigma^2\right),
\end{align}
where the positive scalar $\rho$ characterizes the separation between two cluster means $\mu^*(1+\rho)$ and $\mu^*(1-\rho)$. The standard univariate Gaussian approach fits the data with the model
\begin{align}
\label{eq:normal_dist}
\mathbb{P}_\mu=\mathcal{N}\left(\mu, \sigma^2\right).
\end{align}
In an ideal scenario with an infinite number of samples, the best fit with the univariate approach would be $\bar{\mu} \in \arg \min _{\mu} \operatorname{KL}\left(\mathbb{P}_*, \mathbb{P}_\mu\right)$ where $\operatorname{KL}$ denotes the KL divergence measure. In the following position, we adapt the existing technique on mis-specified GMM \citep{dwivedi2018theoretical} to quantify the bias of such a best fit $\bar{\mu}$ and the true parameter $\mu^{*}$.

\begin{proposition}
\label{prop:misspecify}
   Given the true model (\ref{eq:true_model}) and any $\rho>0$ and $\sigma > 0$, we have
\begin{align}
\label{eq:bound}
\left|\mu^*-\bar{\mu}\right|\leq \rho\left|\mu^*\right|+c\left(\frac{\rho\left|\mu^*\right|}{\sigma}\right)^{1 / 4},
\end{align}
where $c$ is a universal positive constant.   
 \end{proposition}
The proof of the proposition utilizes several bounds between KL divergence, Hellinger distance, and Wasserstein distance and is deferred to the Appendix. 
Bound (\ref{eq:bound}) quantifies the bias of the parameter of the best fitted model from class (\ref{eq:normal_dist}) \citep{dwivedi2018theoretical}. Proposition \ref{prop:misspecify} shows that under such a misspecification scenario, there could be a non-negligible bias of the traditional univariate approach in estimating the anomaly score distribution, and such a bias grows with the separation degree $\rho$. On the other hand, it's known that a properly specified GMM model could consistently approach the true parameter $\mu^*$, resulting in an improvement in error quantification \citep{dwivedi2018theoretical}.

\subsection{Improvement in Score Aggregation} 
In M\textsuperscript{2}AD, we fit a Gamma distribution instead of the traditional chi-square distribution to the aggregated test statistic \(S_{t}\) to account for dependencies among the p-values. We now explore how this relaxation from the chi-square to the Gamma distribution leads to better detection accuracy. To formalize this, we examine the asymptotic behavior of the p-values as the test statistic grows large:

\begin{proposition}
\label{prop:gamma}
Suppose the true distribution of the test statistic \(S_t\) follows a Gamma distribution $\Gamma(\alpha, \theta)$. When using a chi-square distribution \(\chi^2(2 \alpha)\) to approximate the p-value, the ratio of the correct p-value to the incorrect p-value satisfies:
\[
\lim_{S_t \to \infty} \frac{p_{\text{correct}}}{p_{\text{wrong}}} = 
\begin{cases}
1 & \text{if } \theta = 2, \\
\infty & \text{if } \theta < 2, \\
0 & \text{if } \theta > 2.
\end{cases}
\]
\end{proposition}


This theorem confirms that using the chi-square distribution for dependent test statistics can lead to an incorrect p-value calibration. Specifically, when \(\theta < 2\), the p-value under the chi-square assumption becomes disproportionately smaller than the correct Gamma-derived p-value, resulting in overly aggressive detections and a higher risk of false positives. On the other hand, when \(\theta > 2\), the correct p-value can be much smaller, making the test overly conservative and increasing the risk of false negatives, potentially missing critical anomalous behaviors. Adopting the Gamma distribution for score aggregation in dependent scenarios mitigates these issues, leading to more accurate score aggregation.



\section{Experimental Results}
\label{sec:evalution}

\begin{table}[t]
    \centering
    \caption{Public Datasets Summary. The table presents the size of the dataset and its dimension, the number of known anomalies, and the average signal length. These datasets are publicly accessible.}
    \label{tab:public_datasets}
    \resizebox{\linewidth}{!}{%
    \begin{tabular}{lcccc}
    \toprule
    Dataset                   & \# Assets & Dimension   & \# Anomalies   & Avg. Signal           \\
    \midrule
    MSL                       & 27        & 55    & 36                & 4890.59               \\
    SMAP                      & 54        & 25    & 67                & 10618.86              \\
    SMD                       & 28        & 38    & 327               & 50600.89              \\
    \bottomrule
    \end{tabular}}\vspace{-1em}
\end{table}

In this section, we assess the performance of \framework and conduct studies to verify our design choices.

\begin{table*}[!t]
    \centering
    \caption{Benchmark Results. The table summarizes $F1$ and $F_{0.5}$ scores across MSL, SMAP, and SMD datasets. Each model is executed for 5 iterations and we record the mean score and standard deviation.}
    \resizebox{\linewidth}{!}{%
    \begin{tabular}{l*{8}{c}}
    \toprule
               & \multicolumn{2}{c}{MSL} & \multicolumn{2}{c}{SMAP} & \multicolumn{2}{c}{SMD} & \multicolumn{2}{c}{Average} \\
               \cmidrule(lr){2-3}\cmidrule(lr){4-5}\cmidrule(lr){6-7}\cmidrule(lr){8-9}
    Model      & $F1$              & $F_{0.5}$         & $F1$              & $F_{0.5}$    & $F1$              & $F_{0.5}$    & $F1$              & $F_{0.5}$   \\
    \midrule
    \framework              &  \textbf{0.749 $\pm$ 0.014} &  \textbf{0.752 $\pm$ 0.018} &  \textbf{0.716 $\pm$ 0.010} &  0.656 $\pm$ 0.012 &  \textbf{0.913 $\pm$ 0.004} & \textbf{0.906 $\pm$ 0.005} & \textbf{0.789 $\pm$ 0.009} & \textbf{0.769 $\pm$ 0.012} \\
    USAD                    &  0.700 $\pm$ 0.037 &  0.631 $\pm$ 0.044 &  0.650 $\pm$ 0.050 &  0.565 $\pm$ 0.057 &  0.608 $\pm$ 0.002 &  0.644 $\pm$ 0.002 & 0.653 $\pm$ 0.030 & 0.613 $\pm$ 0.035 \\
    OA                      &  0.562 $\pm$ 0.035 &  0.457 $\pm$ 0.034 &  0.577 $\pm$ 0.021 &  0.564 $\pm$ 0.031 &  0.576 $\pm$ 0.003 &  0.471 $\pm$ 0.003 & 0.572 $\pm$ 0.020 & 0.497 $\pm$ 0.022 \\
    LSTM-DT                 &  0.483 $\pm$ 0.006 &  0.408 $\pm$ 0.005 &  0.713 $\pm$ 0.008 &  \textbf{0.677 $\pm$ 0.009} &  0.433 $\pm$ 0.008 &  0.394 $\pm$ 0.007 & 0.543 $\pm$ 0.008 & 0.493 $\pm$ 0.007 \\
    LSTM-AE                 &  0.493 $\pm$ 0.023 &  0.490 $\pm$ 0.027 &  0.672 $\pm$ 0.011 &  0.651 $\pm$ 0.016 &  0.402 $\pm$ 0.007 &  0.395 $\pm$ 0.005 & 0.522 $\pm$ 0.014 & 0.512 $\pm$ 0.016 \\
    LSTM-VAE                &  0.492 $\pm$ 0.009 &  0.476 $\pm$ 0.010 &  0.649 $\pm$ 0.005 &  0.615 $\pm$ 0.005 &  0.412 $\pm$ 0.003 &  0.411 $\pm$ 0.002 & 0.518 $\pm$ 0.006 & 0.501 $\pm$ 0.006 \\
    AT                      &  0.391 $\pm$ 0.018 &  0.455 $\pm$ 0.027 &  0.359 $\pm$ 0.014 &  0.387 $\pm$ 0.034 &  0.533 $\pm$ 0.005 &  0.470 $\pm$ 0.009 & 0.428 $\pm$ 0.012 & 0.437 $\pm$ 0.023 \\
    TimesNet                &  0.276 $\pm$ 0.017 &  0.208 $\pm$ 0.014 &  0.169 $\pm$ 0.008 &  0.117 $\pm$ 0.006 &  0.409 $\pm$ 0.007 &  0.310 $\pm$ 0.006 & 0.285 $\pm$ 0.011 & 0.212 $\pm$ 0.009 \\
    FITS                    &  0.386 $\pm$ 0.005 &  0.338 $\pm$ 0.005 &  0.136 $\pm$ 0.003 &  0.097 $\pm$ 0.002 &  0.316 $\pm$ 0.005 &  0.286 $\pm$ 0.006 & 0.279 $\pm$ 0.004 & 0.240 $\pm$ 0.004 \\
    \bottomrule
    \end{tabular}}\vspace{-1em}
    \label{tab:benchmark}
\end{table*}

\subsection{Evaluation Setup}

\noindent\textbf{Datasets.}
We use three multivariate public datasets with known ground truth anomalies summarized in Table~\ref{tab:public_datasets}. All datasets are pre-split into training and testing sets.
\textbf{Mars Science Laboratory (MSL)} rover dataset is a real public dataset provided by NASA~\citep{lstmdt2018hundman}. This dataset contains 27 assets each has 55 variables; one variable is continuous and the remaining are binary. MSL contains 36 anomalies that were manually labeled by a team of experts.
\textbf{Soil Moisture Active Passive (SMAP)} satellite is another dataset provided by NASA. This dataset has 54 assets with 25 variables each with a similar structure to MSL. 
This dataset has 69 anomalies. Both MSL and SMAP are publicly available~\footnote{\url{https://github.com/khundman/telemanom}}. 
\textbf{Server Machine Dataset (SMD)} is a multivariate public dataset~\footnote{\url{https://github.com/NetManAIOps/OmniAnomaly}} featuring 28 assets collected from a large internet company with 327 anomalies~\citep{smd2019su}. Each asset in this dataset has 38 measurements composing from a variety of continuous and categorical variables. 


\noindent\textbf{Models.}
\label{sec:models}For model comparisons, we evaluate eight models that are competitive to \framework.
\noindent\textbf{LSTM-DT} is a prediction-based model with non-parametric dynamic thresholding developed originally on the NASA datasets~\citep{lstmdt2018hundman}.
\noindent\textbf{LSTM-AE} is an autoencoder architecture with LSTM model that works on reconstructing the signal. When the model fails to reconstruct the signal properly, this indicates the possibility of an anomaly~\citep{malhotra2016lstmautoencoder}.
\noindent\textbf{LSTM-VAE} works on reconstructing time series and utilizes the reconstruction loss to determine if an observation is anomalous.
\noindent\textbf{USAD} is an autoencoder model with two decoders, one of which is adversarially trained. The anomaly score is then constructed from loss values of the weighted sum between the first and second decoder~\citep{usad2020audibert}. 
\noindent\textbf{OmniAnomaly (OA)} is another reconstruction-based model using RNNs with stochastic variable connections to learn the normal pattern of the time series. The reconstruction probability is then used as an anomaly score~\citep{smd2019su}.
\noindent\textbf{AnomalyTransformer (AT)} is a reconstruction-based transformer model that uses attention mechanism to distinguish normal patterns from anomalies~\citep{xu2022anomalytransformer}.
\noindent\textbf{TimesNet} is a 2D variational representation of temporal signals~\citep{wu2023timesnet}.
\noindent\textbf{FITS} is a compact model operating in the frequency domain to learn the amplitude and shift of the time series~\citep{xu2024fits}.
Both TimesNet and FITS leverage reconstruction error to find anomalies.

\noindent\textbf{Evaluation Criteria.}
The nuances in time series data makes point-based evaluations unsuitable~\citep{tatbul2018precisionrecall}. To compute precision and recall metrics, we refer to the commonly used evaluation criteria in \citet{lstmdt2018hundman}, where we assess the performance of the model as to whether or not it was able to locate the anomaly. From the perspective of associates who monitor the data, the desired outcome is to (1) locate true alarms that requires the attention of the associate; and (2) achieve that with very minimal false alarms (false positives). Therefore, we record \textbf{true positive (tp)} if the detected anomaly overlaps with any part of a known ground truth; \textbf{false positive (fp)} if the detected anomaly does not have any counter-part in the ground truth dataset;
and \textbf{false negative (fn)} for all ground truth alarms that were not detected.
Details on the calculation of scores are provided in Appendix~\ref{sec:scoring}.

\subsection{Comparing \framework to Existing Methods.}
Table~\ref{tab:benchmark} showcases the performance of \framework compared to models described earlier in Section~\ref{sec:models}. The table depicts the $F1$ and $F_{0.5}$ scores where we favor precision in the latter.
On average, \framework  achieves the highest scores and outperforms other methods, surpassing the next-best model by 21\% and 25\% in $F_1$ and $F_{0.5}$ scores, respectively. There is a clear performance gain on MSL and SMD datasets. However, we see that LSTM models are competitive in the SMAP dataset.

Competing models suffer in performance on SMD datasets compared to MSL and SMAP, since the number of observations and variable dimension is higher. Recall that MSL and SMAP have only one sensor variable and the remaining are covariates. Moreover, the latter models struggle to score above average when it comes to smaller datasets such as MSL and SMAP.
This suggests that our model is both data-efficient and scalable in contrast to the rest.

\subsection{Sensitivity to error functions.}
The objective of the error function is to unveil the location of the anomalies based on the behavior of the signal and the anomaly type the model is searching for. The selection of this function highly impacts the proceeding steps and the ability to detect anomalies.

\begin{figure}
    \centering
    \includegraphics[width=0.4\linewidth]{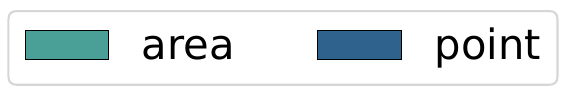}
    \includegraphics[width=0.9\linewidth]{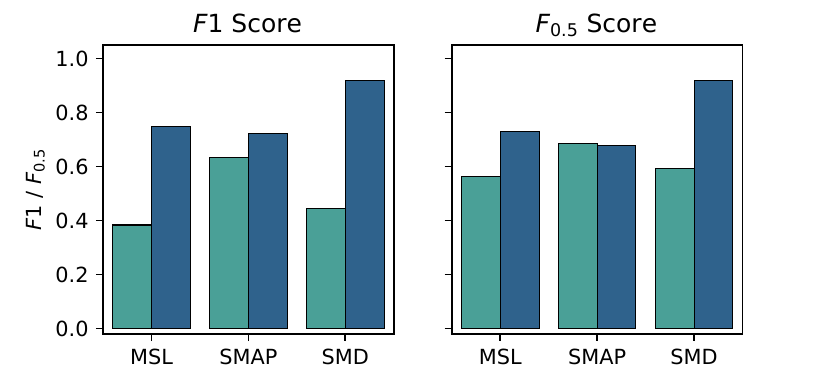}
    \caption{Performance under different error functions on MSL, SMAP, and SMD datasets.}
    \label{fig:error}
\end{figure}

Figure~\ref{fig:error} shows the performance scores under the two error functions suggested in Section~\ref{sec:error}.
For MSL and SMD, the deviation is substantial, specifically in F1 score where we see an improvement by a factor of $\times 2$ when using point errors. As for SMAP, we notice that with area errors we obtain a precision of $0.72$ compared to $0.65$ when using point errors. This emphasizes that depending of the objective criteria, we can favor one discrepancy function over another.

\subsection{Ablation Study}
We perform an ablation study to test the specifications of our anomaly score model under three settings: (1) we apply a static threshold ($\mu \pm 4 \sigma$) on the error values; (2) we compare GMMs to a normal distribution; (3) we compare Gamma to chi-square calibration.
Table~\ref{tab:ablation} presents the results, with \framework configuration shown in the bottom row. First, we see that using Gamma calibration, which properly accounts for the dependency between p-values during score aggregation, leads to substantial gains in both \(F_1\) and \(F_{0.5}\) scores compared to chi-square calibration. Furthermore, when adopting the GMM model for anomaly score aggregation, we observe the best performance across all baselines. 

\begin{table}[t]
    \centering
    \caption{Ablation Study on SMD dataset.}
    \resizebox{\linewidth}{!}{%
    \begin{tabular}{lcc}
    \toprule
    Variation & $F_1$ & $F_{0.5}$ \\
    \midrule
    static                                 & 0.745 $\pm$ 0.000 & 0.724 $\pm$ 0.000 \\
    GMM with $\chi^2$ calibration          & 0.258 $\pm$ 0.032 & 0.426 $\pm$ 0.043 \\
    normal with $\chi^2$ calibration       & 0.317 $\pm$ 0.050 & 0.490 $\pm$ 0.059 \\
    normal with $\Gamma$ calibration       & 0.905 $\pm$ 0.000 & 0.890 $\pm$ 0.001 \\
    \midrule
    GMM with $\Gamma$ calibration          & 0.913 $\pm$ 0.004 & 0.906 $\pm$ 0.005 \\
    \bottomrule
    \end{tabular}}
    \vspace{-1em}
    \label{tab:ablation}
\end{table}

\subsection{Interpretability}
\begin{figure}
    \centering
    \begin{subfigure}{.47\linewidth}
      \centering
      \includegraphics[width=\linewidth]{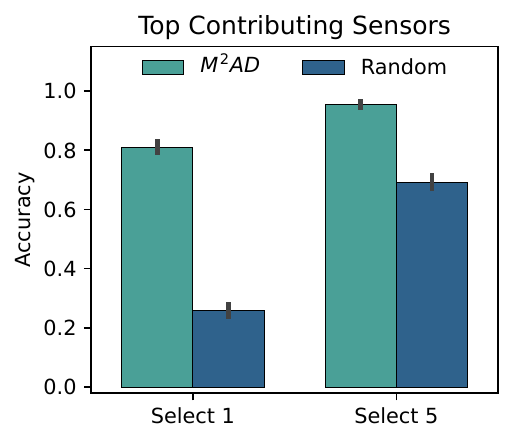}
      \caption{Interpretability.}
      \label{fig:interpretability}
    \end{subfigure}%
    \quad
    \begin{subfigure}{.47\linewidth}
      \centering
      \includegraphics[width=\linewidth]{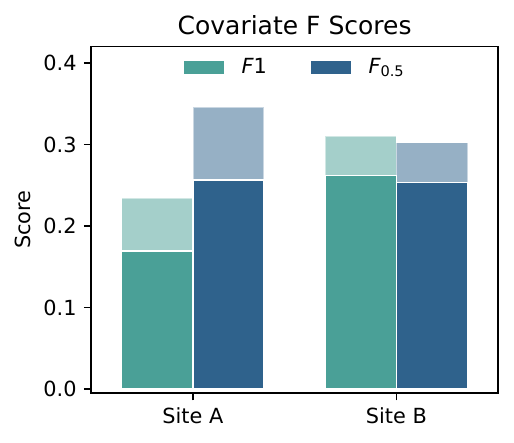}
      \caption{Scores with covariates.}
      \label{fig:cs-performance-covariates}
    \end{subfigure}
    \caption{(a) Accuracy of top contributing sensors to the anomaly score compared to a random selection in SMD. (b) Change in model performance without covariates (dark) and with covariates (light).}
    \label{fig:results-3}
\end{figure}

One of most important aspects in our design is that we can isolate the cause of the anomaly by finding the most contributing sensor to the anomaly score.
Fortunately, the SMD dataset provides an interpretability document that summarizes for each ground truth anomaly, which sensors caused it.
Figure~\ref{fig:interpretability} shows how the top contributing sensors detected by \framework agree with the sensors provided as ground truth in SMD for the correctly identified anomalies (297 out of 327). When we select the top contributing sensors and compare it to randomly selecting one and five sensors, respectively, we find that \framework is correct 81\% of time. Moreover, if we observed the top-5 contributing sensors to the anomaly score, we reach 95\%.

\newcommand{\site}{Site A\xspace}
\newcommand{\ssite}{Site B\xspace}

\section{Case Study: Amazon Fulfillment Centers}
\label{sec:case_study}
The motivation of \framework began to enhance Amazon's predictive maintenance efforts.
At Amazon's fulfillment centers, there are many equipment that are monitored on each individual sensor separately. The systems alert the respective technician when the expected threshold repeatedly exceeds the standard threshold. This process results in a high number of false alarms especially when the sensory data is noisy.
In this section, we present a real-world evaluation of our system on equipment data collected from Amazon fulfillment centers.

\begin{table}[t]
    \centering
    \caption{Data Summary. \site has a larger number of equipment data that we collected compared to \ssite. Moreover, the number of sensors attached to the equipment are higher in \site.}
    \begin{tabular}{lcccc}
    \toprule
            &                   & \multicolumn{3}{c}{dimension} \\
                                \cmidrule(lr){3-5}
    {}      &    \# equipment   & median  &  min & max \\
    \midrule
    \site    &      82           & 51    & 11    & 121  \\
    \ssite    &      48           & 21    & 21    & 191  \\
    \bottomrule
    \end{tabular}
    \label{tab:cs-data-summary2}
\end{table}

\begin{table*}[ht]
    \centering
    \caption{\framework evaluation on Amazon data. The table depicts precision, recall, and $F$ scores under two evaluation strategies. Predictive is a strict strategy where the detection must be at least a day before the work order. Detection is lenient strategy where any detection within a 7-day window counts towards a correct identification.}\vspace{-1em}
    \begin{tabular}{ll*{8}{c}}
    \toprule
                          & & \multicolumn{4}{c}{Predictive}                                  &   \multicolumn{4}{c}{Detection}\\
                          \cmidrule(lr){3-6}\cmidrule(lr){7-10}
    site & work order type &   Precision &  Recall &    $F1$ &   $F_{0.5}$      &  Precision &  Recall &    $F1$ &   $F_{0.5}$ \\
    \midrule
    \site &          total &  0.353 &    0.174 &  0.233 & 0.293       &  0.605 & 0.362 & 0.453 & 0.534 \\
         &     breakdowns &  0.274 &    0.178 &  0.216 & 0.247       &  0.482 & 0.364 & 0.415 & 0.453 \\
    \ssite &          total &  0.269 &    0.448 &  0.336 & 0.293       &  0.369 & 0.632 & 0.466 & 0.403 \\
         &     breakdowns &  0.143 &    0.528 &  0.225 & 0.167       &  0.192 & 0.717 & 0.303 & 0.225 \\
    \bottomrule
    \end{tabular}\vspace{-1em}
    \label{tab:cs-performance}
\end{table*}

\noindent\textbf{Dataset.} 
Our study was performed on Amazon's proprietary data featuring equipment sensors.
We collected 10 months worth of data for two sites \site and \ssite with varying measurements.
These measurements are gathered from two independent systems: amperage which summarizes amperage from VFD motors, and Monitron which collects vibration and temperature measurements.
In addition, we collect status variables to use as covariates.
Covariates capture non-physical contextual information that helps us model system behavior. For example, we include throughput, which is the number of processed packages, after discretizing it into low, medium, and high bins.
Moreover, we use the percentage of time where the machine is available on planned production time as equipment availability.
Each equipment has a varying number of sensors that monitors its condition. 
In other words, the dimension of the data varies by asset.
Table~\ref{tab:cs-data-summary2} shows a summary of the data used.

Like most industrial datasets, this dataset does not have exact labels on whether or not an anomaly occurred.
To measure the performance of our model, we use filed work orders as a proxy to know if the asset has been investigated.
We look at work orders from two perspective: \textit{breakdown} work orders which describes if the equipment broke down and needed attention and action to be taken by a technician; and the \textit{total} work orders that had caused downtime to the equipment (including breakdowns) and needed technician intervention for unplanned maintenance.

\noindent\textbf{Baseline.}
The current production model at Amazon is based on a combination of statistical and ISO thresholds for vibration sensors only.
Out of all the raised alarms in Sites A and B, only 19.2\% and 23.1\% of them are correct, respectively.
More details on the baseline performance are provided in Appendix~\ref{sec:case_study_baseline}.

\begin{figure}
    \centering
    \includegraphics[width=0.9\linewidth]{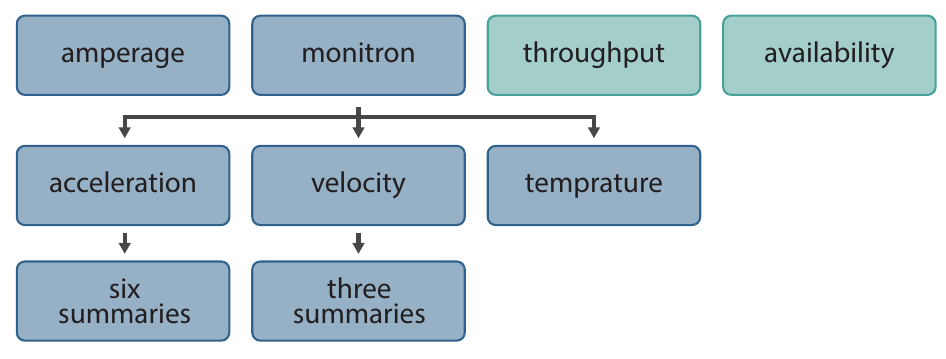}
    \caption{Hierarchical structure of data sources. We collect data from amperage and Monitron which contain sensory information (blue). We also leverage the throughput and availability as covariates (green).}
    \label{fig:hierarchy}
\end{figure}

\noindent\textbf{Experiment Setup.}
We utilize hourly data with median aggregations and select the hyperparameters of the model based on the nature of the problem and dataset.
For Monitron sensors, we set the number of components for the GMM as $m_k=1$. For amperage, we set the number of components to $m_k=2$. Having two components allowed us to fit both ``operation'' mode and ``shutdown'' mode better \citep{he2024wmad}.

Based on the hierarchy organization of the data shown in Figure~\ref{fig:hierarchy}, and that one data source (Monitron) out-numbers the number of sensors in the amperage system, we construct an equally divided weighing vector:  $\lambda_k = 1 / (\text{\# systems} \cdot \text{\# sensors} \cdot \text{\# summaries})$. 
This ensures that under-represented sensors, e.g. amperage, does not lose its contribution to the anomaly score. 

\noindent\textbf{Predictive Performance.}
Table~\ref{tab:cs-performance} depicts the performance of \framework on the two sites.
On the left-hand side, we view the performance under a predictive setting. This means that we only count a detection as true positive if it happens between a day to 7 days \emph{before} a work order is initiated. The assumption here is that a precursor to the actual breakdown can be detected within the monitored time series.
Compared to the baseline model, \framework improves precision by 16\% in Site A and 4\% in Site B.

We also demonstrate the performance of the framework using the detection strategy, this follows the regular evaluation criteria introduced in Section~\ref{sec:evalution}. 
Naturally, we observe a higher performance in the detection evaluation strategy compared to the predictive one. 

\noindent\textbf{Importance of Covariates.}
Figure~\ref{fig:cs-performance-covariates} showcases the model performance with and without including ``throughput'' and ``availability'' as covariates to the model. 
There is a clear improvement on all fronts after adding covariates to the model. Specifically, the number of false positives reduced by 16.5\% in \site and 12\% in \ssite on average for both work order types. 
Note that the scores shown in Figure~\ref{fig:cs-performance-covariates} differ from Table~\ref{tab:cs-performance} since the subset of assets and work orders has changed. This stems from the data collection process where not all asset have an associated throughput and availability measurements. Only 46\% and 50\% of assets have covariates in \site and \ssite respectively.

\noindent\textbf{Discussion.}
In \site, we observe high precision and low recall scores while \ssite exhibits counter scores. This traces back to number of \textit{breakdown} work orders created in \site, which are $6\times$ the order of \ssite, making it more likely to find anomalies.
On average, the model predicts one anomaly every 2 months per asset. Having a low number of flagged anomalies is crucial to gain the trust of on-site technicians.

The flexibility of \framework has been key when certain sensor data becomes unavailable. Since the framework design is at sensor-level, we simply omit the contribution of that sensor.
However, limitations of our model arise when the training distribution completely shifts from the target distribution, necessitating model retraining.
\section{Conclusion}
\label{sec:conclusion}

In this paper, we propose \framework, a modular framework for detecting anomalies in multi-sensor multi-system time series data. We learn the temporal patterns and cross-sensor dependencies using a forecasting model which then feeds to a discrepancy function to capture the error vectors. The model uses a data-driven approach using GMMs and Fisher's method to construct a global anomaly score for each observation in the time series dataset. 

We show the effectiveness of our model by comparing it to 8 competitive methods on three multivariate public datasets. Moreover, we highlight our case study in Amazon fulfillment centers in utilizing the proposed framework, where it improved on the current baseline precision by 16\%. \framework's flexibility allows it to function even when some sensors are unavailable, enhancing its practicality for deployment.



\bibliographystyle{apalike}
\bibliography{ref}

\clearpage
\newpage


%
%





%

%

\appendix
\onecolumn
\aistatstitle{Supplementary Materials for \framework: Multi-Sensor Multi-System Anomaly Detection using Global Scoring and Calibrated Thresholding}

In this supplementary material, we provide more details on the framework as well as additional experiments. Additionally, we illustrate the limitations of our methodology.

\section{Additional \framework Details}
\subsection{Problem Statement}

Given $\mathcal{S}$ systems, where we observe a multivariate time series $\mathcal{X}_s = \{x_{s_1}, \dots, x_{s_T}\}$ for all $s \in \mathcal{S}$ where \(x_{s_t} \in \mathbb{R}^{d_s}\) and system $s$ has $d_s$ sensors, and $T$ is the length of the time series, we want to find the anomalies present in $\mathcal{X}_s$ for all $s \in \mathcal{S}$.

For simplicity of notation, we denote the combined observations as a single multivariate time series \(\mathcal{X} = \{x_1, \dots, x_T\}\), where \(x_t \in \mathbb{R}^d\) and \(d\) represents the total number of sensors. Optionally, we have covariates \(\mathcal{C} = \{c_1, \dots, c_T\}\) which is a categorical time series representing the status of the asset.

Below is the expected input of \framework
\[
\mathcal{D} = 
\left[
\begin{array}{cccc|ccc}
    x_{11} & x_{12} & \dots  & x_{1d} & c_{11}   & \dots  & c_{1m}\\
    x_{21} & x_{22} & \dots  & x_{2d} & c_{21}   & \dots  & c_{2m}\\
    \vdots & \vdots & \ddots & \vdots & \vdots  & \ddots & \vdots\\
    x_{T1} & x_{T2} & \dots  & x_{Td} & c_{T1}   & \dots  & c_{Tm}
\end{array}
\right]_{T\times (d + m)}
\]

where $T$ is the length of the time series, $d$ is the number of sensors we have in total, and $m$ is the number of status time series we include as covariates in the model. 

Our goal is to identify anomalous regions by finding $\mathcal{A} = \{t: z_t = 1\}$ in a given asset where $z_t \in \{0, 1\}$. Note that we do not have any prior information on the existence of anomalies, in fact $\mathcal{A}$ is often an empty set.

The model works in two phases: training and inference.
We split the data into $\mathcal{D}_{train} = \{(x_t, c_t)_{i=1}^n\}$ and $\mathcal{D}_{test} = \{(x_t, c_t)_{i=n}^T\}$ where $n < T$ and we will use $\mathcal{D}_{train}$ and $\mathcal{D}_{test}$ train our model and test it respectively.
The main assumption of the model is that both $\mathcal{D}_{train}$ and $\mathcal{D}_{test}$ are operating under the same condition. Moreover, $\mathcal{D}_{train}$ does not contain anomalies such that we learn a proper model of normal data.

\subsection{Time Series Prediction}
In order to feed the time series into an LSTM model, we use a sliding window approach to slice the data. We select a window size $w$, we find a corresponding slice which gives us $x_{w_t}$ and $c_{w_t}$ starting at time $t$.
\begin{align*}
    x_{w_t} &= \{x_t, x_{t-1}, \dots, x_{t-w}\} \\
    c_{w_t} &= \{c_t, c_{t-1}, \dots, c_{t-w}\}
\end{align*}
where we now have $y_{t+1} = f(x_{w_t}, c_{w_t})$. We find $y_t$ for all $t \in [w+1, n]$ in the training phase.

\subsection{Discrepancy Calculation}
With absolute point-wise difference error, we observe many spikes. In an effort to reduce extreme values, we apply an exponentially-weighted average (EWMA) to produce smoothed errors~\citep{hunter1986exponentially, lstmdt2018hundman}.

\subsection{Choice of Number of Components in GMM}
\label{choice_of_mk}
In practice, the number of components \( m_k \) is determined by the operational characteristics of each sensor system. For example, amperage sensors, which frequently undergo shutdowns and restarts \citep{he2024wmad}, often require a Gaussian mixture model (GMM) with two components (\( m_k = 2 \)), while more stable systems, such as Monitron, can typically be modeled with a single component (\( m_k = 1 \)).  

When such prior knowledge is unavailable, we determine \( m_k \) using the Bayesian Information Criterion (BIC):  
\begin{align*}
\operatorname{BIC} = -2 \cdot \text{Likelihood} + k \log (n),
\end{align*}
where \( k \) is the number of parameters and \( n \) is the sample size. BIC balances model complexity and goodness of fit, providing a principled approach for selecting the optimal number of components.


\section{Proof of Propositions}
\subsection{Proof of Proposition \ref{prop:misspecify}}

To prove the proposition, we follow the similar arguments as in \cite{dwivedi2018theoretical}. We first establish an upper bound on the Wasserstein distance between the true distribution and the best fit model from the misspecified distribution, then proceed to connect such a Wasserstein distance with the bias term $\left|\bar{\mu}-\mu^{*}\right|$.

We first introduce the notation for the measure of true distribution (\ref{eq:true_model}) and the misspecified distribution (\ref{eq:normal_dist}):
$$
G_{*}=\frac{1}{2} \delta_{\mu^{*}(1+\rho)}+\frac{1}{2} \delta_{\mu^{*}(1-\rho)} \text {, and } G(\mu)=\delta_{\mu}.
$$

where $\delta_{\mu}$ denotes the dirac measure at $\mu$. Here $G_{*}$ is the mixing measure of interest and $G(\mu)$ is a measure with a (misspecified) univariate Gaussian distribution. Since the true model does not belong to the univariate Gaussian distribution class, the ``best fit" would be the projection parameter 
$$
\bar{\mu} \in \underset{\mu}{\operatorname{argmin}}  \operatorname{KL}\left(P_{G_{*}}, P_{G(\mu)}\right).
$$
where $\operatorname{KL}(\mathbb{P}, \mathbb{Q})=\int p(x) \log \frac{p(x)}{q(x)} d x$ is the Kullback-Leibler divergence between two distribution $\mathbb{P}$ and $\mathbb{Q}$. Further denote $W_{2}^{2}(\mathbb{P}, \mathbb{Q})$ as the second-order Wasserstein distance between $\mathbb{P}$ and $\mathbb{Q}$ \citep{villani2009optimal}, applying the arguments from (26) to (29) in \cite{dwivedi2018theoretical} yields
\begin{align}
\label{eq:distance_inequality}
2 \sigma \sqrt{C} W_{2}^{4}\left(G_{*}, G(\bar{\mu})\right) \leqslant W_{2}\left(G_{*}, G\left(\mu^{*}\right)\right)
\end{align}

where $C$ is a universal constant and $\sigma$ is the standard deviation defined in (\ref{eq:true_model}). Note that $G\left(\mu^{*}\right) \neq G_{*}$ as they belong to different distribution classes. Next, we establish an upper bound for the distance $W_{2}\left(G_{x}, G(\bar{\mu})\right)$ by deriving an upper bound for the distance $W_{2}\left(G_{x}, G\left(\mu^{*}\right)\right)$. According to variational formulation in (24d) of \cite{dwivedi2018theoretical}, we have that for the pair $(G^*, G(\mu^{*}))$, we can construct the coupling matrix $T$ and the corresponding distance matrix $D$ as
$$
T=\left[\begin{array}{c}
\frac{1}{2} \\
\frac{1}{2}
\end{array}\right], \quad D=\left[\begin{array}{l}
\left(\rho \mu^{*}\right)^{2} \\
\left(-\rho \mu^{*}\right)^{2}
\end{array}\right],
$$
and further obtain
\begin{align}
\label{eq:variational_upper_bound}
W_{2}^{2}\left(G_{*}, G\left(\mu^{*}\right)\right) \leqslant \sum_{i=1}^{2} \sum_{j=1}^{2} T_{i j} D_{i j}=\rho^{2} \mu^{*^{2}}.
\end{align}
Combining (\ref{eq:variational_upper_bound}) with (\ref{eq:distance_inequality}) implies
\begin{align}
\label{eq:upper_bound_on_W_distance}
W_{2}\left(G_{*}, G(\bar{\mu})\right) \leqslant c\left(\frac{\rho\left|\mu^{*}\right|}{\sigma}\right)^{1 / 4}
\end{align}
with $c=1 /(4 C)^{1 / 8}$. We then apply the variation formulation (4d) in \cite{dwivedi2018theoretical} again but this time w.r.t. the pair $\left(G_{*}, G(\bar{\mu})\right)$ with
$$
T^{\prime}=T=\left[\begin{array}{c}
\frac{1}{2} \\
\frac{1}{2}
\end{array}\right] \quad D^{\prime}=\left[\begin{array}{c}
{\left[(1-\rho) \mu^{*}-\bar{\mu}\right]^{2}} \\
{\left[(1+\rho) \mu^{*}-\bar{\mu}\right]^{2}}
\end{array}\right],
$$
we then have
\begin{align}
W_{2}\left(G_{*}, G(\bar{\mu})\right) \geqslant \min _{i, j} \sqrt{D_{i, j}^{\prime}} & =\min \left\{(1-\rho) \mu^{*}-\bar{\mu},(1+\rho) \mu^{*}-\bar{\mu}\right\} \\
& \geqslant |\mu^{*}-\bar{\mu}|-\rho\left|\mu^{*}\right|,
\end{align}

this together with (\ref{eq:upper_bound_on_W_distance}) leads to the results in Proposition \ref{prop:misspecify}.

\subsection{Proof of Proposition \ref{prop:gamma}}

Denote $F_{S}(s ; \alpha, \theta)$ as the CDF of the Gamma distribution $\Gamma(\alpha, \theta)$ and $F_{S}(s; \alpha, 2)$ as the CDF of the $\chi^{2}(2 \alpha)$ distribution (as $\chi^{2}(2 \alpha)$ is equivalent to $\Gamma(\alpha, 2)$). Notice that here we neglect the subscript $t$ for notation simplicity. We aim to analyze the asymptotic behavior of
$$
\frac{p_{\text {correct }}}{p_{\text {wrong }}}=\frac{1-F_{S}(s; \alpha, \theta)}{1-F_{S}(s ; \alpha, 2)}.
$$
as $s \rightarrow \infty$. Notice that the PDF of $F_S(S;\alpha, \theta)$ is $ f_{S}(s ; \alpha, \theta):=\frac{s^{\alpha-1} \exp (-s / \theta)}{\Gamma(\alpha) \theta^{\alpha}}$ with $\Gamma(\alpha)$ being the Gamma function. Therefore, by L'Hospital's Rule we have
$$
\lim _{s \rightarrow \infty} \frac{p_{\text {correct }}}{p_{\text {wrong }}}=\lim _{s \rightarrow \infty} \frac{f_{S}(s ; \alpha, \theta)}{f_{S}(s ; \alpha, 2)}=\frac{2^{\alpha}}{\theta^{\alpha}} \lim _{s \rightarrow \infty} \exp \left(-s\left(\frac{1}{\theta}-\frac{1}{2}\right)\right) \text {. }
$$
which leads to the result in Proposition \ref{prop:gamma}.

\section{Experiments}
In this section we provide more details on the experimental setup as well as some addition results.

\subsection{Models}
We compare our model to eight state-of-the-art machine learning models for unsupervised time series anomaly detection.
These models were chosen based on their similarity in nature to our proposed model.

\noindent\textbf{LSTM-DT} is a prediction-based model that uses LSTM to forecast a time series, then applies dynamic thresholding on the absolute point-wise errors computed from the original signal and forecasted one~\citep{lstmdt2018hundman}. LSTM-DT uses a window size of 250 based on recommendations by the original authors.

\noindent\textbf{LSTM-AE} is an autoencoder architecture with LSTM model that works on reconstructing the signal. When the model fails to reconstruct the signal properly, this indicates the possibility of an anomaly~\citep{malhotra2016lstmautoencoder}. We set the window size to 100 since it is a reconstruction based model.

\noindent\textbf{LSTM-VAE} Similar to LSTM-AE, LSTM-VAE works on reconstructing time series using a variational autoencoder. The reconstruction error is utilized to locate anomalies.

\noindent\textbf{USAD} is an autoencoder model with two decoders, one of which is adversarially trained. The anomaly score is then constructed from a weighted sum between the loss first and second decoder~\citep{usad2020audibert}. USAD does not include a thresholding technique, therefore, we experimented with multiple values and found that 98\textsuperscript{th} percentile yielded the best results.

\noindent\textbf{OmniAnomaly (OA)} is another reconstruction based model using RNNs with stochastic variable connections to learn the normal pattern of the time series. The reconstruction probability is then used as an anomaly score~\citep{smd2019su}.

\noindent\textbf{AnomalyTransformer (AT)} is a transformer model that uses attention mechanism to compute the association discrepancy, which distinguishes normal patterns from anomalies. They use a minimax training strategy to amplify the dichotomy between normal and abnormal patterns~\citep{xu2022anomalytransformer}.

\noindent\textbf{TimesNet} is a novel framework which transforms 1D temporal signals into a 2D representation consisting of intraperiod- and interperiod-variations. Their framework can be used for multiple tasks, specifically for anomaly detection, they employ reconstruction error~\citep{wu2023timesnet}.

\noindent\textbf{FITS} is a compact neural network model for time series analysis operating in the frequency domain. The model aims to learn the amplitude and shift of the time series by interpolating the frequency representation. They leverage reconstruction error for testing the anomaly detection task~\citep{xu2024fits}.

To run these models, we used the code provided directly by~\url{https://github.com/sintel-dev/Orion}~\citep{alnegheimish2022sintel}, \url{https://github.com/manigalati/usad}~\citep{usad2020audibert}, \url{https://github.com/NetManAIOps/OmniAnomaly}~\citep{smd2019su}, \url{https://github.com/thuml/Time-Series-Library}~\citep{wu2023timesnet}, and \url{https://github.com/VEWOXIC/FITS}~\citep{xu2024fits}.

\subsection{Scoring}
\label{sec:scoring}
We compute how many true positives, false positives, and false negatives we obtain on a signal level. Because often times the model fails to detect any ground truth anomalies, the number of true positives is zero, making precision and recall scores undefined. To aggregate the results in a meaningful approach, we compute the performance scores with respect to the dataset $D$.
\[
\text{precision}_D = \frac{\sum_{i=1}^{|D|} TP_i}{\sum_{i=1}^{|D|} TP_i + FP_i}
\quad\quad
\text{recall}_D = \frac{\sum_{i=1}^{|D|} TP_i}{\sum_{i=1}^{|D|} TP_i + FN_i}
\]
where $i$ denotes one signal from the dataset $D$.

Similarly, we compute the $F$ scores in a similar fashion

\[
F_{1_D} = 2 \cdot \frac{\text{precision}_D \cdot \text{recall}_D}{\text{precision}_D + \text{recall}_D}
\quad\quad
F_{0.5_D} = (1 + 0.5^2) \cdot \frac{\text{precision}_D \cdot \text{recall}_D}{0.5^2 \cdot \text{precision}_D + \text{recall}_D}
\]

\subsection{Number of Components in GMM}
Using the criterion described in Appendix~\ref{choice_of_mk}, we conducted an exploratory experiment to calculate the BIC scores for models with varying numbers of components on benchmark datasets (see Table~\ref{tab:bic_scores}). Notably, we observed that beyond three components, the improvement in BIC scores diminishes, particularly on the SMD dataset. This observation aligns with our choice of components and shows how to use the BIC criterion as a principled method for determining the number of components.

\begin{table}[ht]
    \centering
    \caption{BIC values across datasets with different number of GMM components.}
    \begin{tabular}{lccc}
    \toprule
        \# components    & MSL & SMAP    & SMD \\
        \midrule
        1   & -11783.36    & -15964.53     & -166770.01\\
        2   & -12885.25    & -17811.11     & -190736.41\\
        3   & -13168.31    & -17955.04     & -193107.46\\
        4   & -13012.05    & -17853.51     & -193909.34\\
        5   & -13213.42    & -17976.60     & -194323.92\\
    \bottomrule
    \end{tabular}
    \label{tab:bic_scores}
\end{table}

\subsection{Case Study}
\noindent\textbf{Baseline.}
\label{sec:case_study_baseline}
The table below shows the baseline performance of the current condition-based monitoring model at Amazon sites. The model is based on a combination of statistical and ISO thresholds for vibration sensors in Monitron data. However, the existing model in production cannot be easily extended to include more sensors. Our experiments show up to 16\% improvement in precision and 15\% improvement in recall in Site A over the current baseline model, and an increase of 4\% in precision and 22\% in recall scores in Site B.

\begin{table}[ht]
    \centering
    \caption{Baseline performance compared to \framework on Site A and Site B.}
    \begin{tabular}{lcccc}
    \toprule
                    & \multicolumn{2}{c}{Baseline} & \multicolumn{2}{c}{\framework} \\
                    \cmidrule(lr){2-3}\cmidrule(lr){4-5}
                    & Precision & Recall    & Precision & Recall \\
                    \midrule
         Sita A     & 19.2\%   & 2.7\%     & 35.3\%    & 17.4\% \\
         Site B     & 23.1\%    & 22.5\%    & 26.9\%    & 44.8\% \\
    \bottomrule
    \end{tabular}
    \label{tab:baseline_performance}
\end{table}

\noindent\textbf{Experiment Setup.}
We apply a categorical transformation to availability and throughput. Availability is the percentage of time where the machine is available on planned production time, we convert this percentage to a binary value where $> 98\%$ indicates that the asset is available.
As for throughput, which measures how many packages have been processed, is an asset-specific measure which we categorize into: low, medium, high based on quantile values of the training data.
With respect to pre-processing, first the data is resampled to an hourly-level with the median aggregation. In addition, we scale the data into the range (-1, 1) and interpolate missing values.
We select the hyperparameters of the model based on the nature of the problem and dataset.
We choose a window size of 120, which is equivalent to 5-day span of historical values. Moreover, we use an area-error with $l=2$, a small $l$ is chosen as it determines the lag time needed before being able to execute the model. In our setting, we prioritize an earlier detection of anomalies.
We train the LSTM model for 30 epochs and early stopping, which was enough for the model to converge in our preliminary experiments.
For monitron sensors, we set the number of components for the GMM as $m_k=1$. For amperage, we set the number of components to $m_k=2$, having two components allowed us to fit the amperage errors better. More specifically, the pattern of amperage data changes between ``operation'' mode and ``shutdown'' mode, which exhibits a bimodal distribution.

Based on the hierarchy organization of the data show in Figure~\ref{fig:hierarchy}, and that one data source (monitron) out-numbers the number of sensors in the amperage system, we construct an equally divided weighing vector:  
\[
\lambda_k = \frac{1}{\text{\# systems} \cdot \text{\# sensors} \cdot \text{\# summaries}}
\]
For example, each measurement in acceleration will have: $\lambda_{\text{acceleration}} = 1 / (2 \times 3 \times 6)$.
This ensures that under-represented sensors, e.g. amperage, does not lose its contribution to the anomaly score.

\subsection{Benchmark}
We run the benchmark on a total of 109 signals from MSL, SMAP, and SMD dataset. Table~\ref{tab:full-benchmark} shows the overall performance for the datasets used.

\begin{table}[t]
    \centering
    \caption{Benchmark Results. The table summarizes precision (pre), recall (rec), $F_1$ and $F_{0.5}$ scores across NASA dataset. Each model is executed for 5 times and we record the mean score and standard deviation}
\begin{tabular}{ll*{4}{c}}
\toprule
Model &     & $F_1$ & $F_{0.5}$ & Precision & Recall \\
\midrule
\multirow{3}{*}{\framework}
            & MSL   & 0.749 $\pm$ 0.032 & 0.752 $\pm$ 0.040 &     0.754 $\pm$ 0.046 &  0.744 $\pm$ 0.023 \\
            & SMAP  & 0.716 $\pm$ 0.021 & 0.656 $\pm$ 0.028 &     0.622 $\pm$ 0.031 &  0.845 $\pm$ 0.013 \\
            & SMD   & 0.913 $\pm$ 0.004 & 0.906 $\pm$ 0.005 &     0.904 $\pm$ 0.004 &  0.926 $\pm$ 0.001 \\
\midrule
\multirow{3}{*}{USAD} 
            & MSL   & 0.700 $\pm$ 0.083 & 0.631 $\pm$ 0.098 &     0.593 $\pm$ 0.104 &  0.861 $\pm$ 0.028 \\
            & SMAP  & 0.650 $\pm$ 0.111 & 0.565 $\pm$ 0.128 &     0.521 $\pm$ 0.133 &  0.893 $\pm$ 0.012 \\
            & SMD   & 0.608 $\pm$ 0.004 & 0.644 $\pm$ 0.005 &     0.670 $\pm$ 0.007 &  0.557 $\pm$ 0.005 \\
\midrule
\multirow{3}{*}{OA} 
            & MSL   & 0.562 $\pm$ 0.079 & 0.457 $\pm$ 0.076 &     0.406 $\pm$ 0.072 &  0.922 $\pm$ 0.036 \\
            & SMAP  & 0.577 $\pm$ 0.047 & 0.564 $\pm$ 0.069 &     0.557 $\pm$ 0.083 &  0.603 $\pm$ 0.008 \\
            & SMD   & 0.576 $\pm$ 0.006 & 0.471 $\pm$ 0.006 &     0.419 $\pm$ 0.006 &  0.923 $\pm$ 0.002 \\
\midrule
\multirow{3}{*}{LSTM-DT} 
            & MSL   & 0.483 $\pm$ 0.014 & 0.408 $\pm$ 0.012 &     0.370 $\pm$ 0.011 &  0.694 $\pm$ 0.020 \\
            & SMAP  & 0.713 $\pm$ 0.018 & 0.677 $\pm$ 0.021 &     0.655 $\pm$ 0.023 &  0.782 $\pm$ 0.017 \\
            & SMD   & 0.433 $\pm$ 0.019 & 0.394 $\pm$ 0.015 &     0.373 $\pm$ 0.020 &  0.523 $\pm$ 0.073 \\
\midrule
\multirow{3}{*}{LSTM-AE} 
            & MSL   & 0.493 $\pm$ 0.052 & 0.490 $\pm$ 0.061 &     0.489 $\pm$ 0.067 &  0.500 $\pm$ 0.044 \\
            & SMAP  & 0.672 $\pm$ 0.025 & 0.651 $\pm$ 0.035 &     0.638 $\pm$ 0.041 &  0.710 $\pm$ 0.013 \\
            & SMD   & 0.402 $\pm$ 0.015 & 0.395 $\pm$ 0.012 &     0.392 $\pm$ 0.011 &  0.412 $\pm$ 0.020 \\
\midrule
\multirow{3}{*}{LSTM-VAE} 
            & MSL   & 0.492 $\pm$ 0.020 & 0.476 $\pm$ 0.021 &     0.466 $\pm$ 0.023 &  0.522 $\pm$ 0.023 \\
            & SMAP  & 0.649 $\pm$ 0.011 & 0.615 $\pm$ 0.012 &     0.595 $\pm$ 0.013 &  0.713 $\pm$ 0.016 \\
            & SMD   & 0.412 $\pm$ 0.008 & 0.411 $\pm$ 0.005 &     0.410 $\pm$ 0.007 &  0.415 $\pm$ 0.019 \\
\midrule
\multirow{3}{*}{AT} 
            & MSL   & 0.391 $\pm$ 0.039 & 0.455 $\pm$ 0.061 &     0.516 $\pm$ 0.095 &  0.322 $\pm$ 0.050 \\
            & SMAP  & 0.359 $\pm$ 0.031 & 0.387 $\pm$ 0.076 &     0.415 $\pm$ 0.121 &  0.334 $\pm$ 0.044 \\
            & SMD   & 0.533 $\pm$ 0.012 & 0.470 $\pm$ 0.019 &     0.436 $\pm$ 0.023 &  0.689 $\pm$ 0.022 \\
\midrule
\multirow{3}{*}{TimesNet} 
            & MSL   & 0.276 $\pm$ 0.038 & 0.208 $\pm$ 0.031 &     0.179 $\pm$ 0.027 &  0.611 $\pm$ 0.071 \\
            & SMAP  & 0.169 $\pm$ 0.018 & 0.117 $\pm$ 0.014 &     0.098 $\pm$ 0.012 &  0.630 $\pm$ 0.036 \\
            & SMD   & 0.409 $\pm$ 0.015 & 0.310 $\pm$ 0.013 &     0.266 $\pm$ 0.012 &  0.887 $\pm$ 0.028 \\
\midrule
\multirow{3}{*}{FITS} 
            & MSL   & 0.386 $\pm$ 0.011 & 0.338 $\pm$ 0.010 &     0.312 $\pm$ 0.010 &  0.506 $\pm$ 0.012 \\
            & SMAP  & 0.136 $\pm$ 0.007 & 0.097 $\pm$ 0.005 &     0.082 $\pm$ 0.004 &  0.397 $\pm$ 0.020 \\
            & SMD   & 0.316 $\pm$ 0.012 & 0.286 $\pm$ 0.013 &     0.269 $\pm$ 0.013 &  0.383 $\pm$ 0.011 \\
\bottomrule
\end{tabular}
    \label{tab:full-benchmark}
\end{table}

\newpage

\begin{wrapfigure}{r}{0.5\textwidth}
    \vspace{-1cm}
    \centering
    \begin{subfigure}{.47\linewidth}
      \centering
      \includegraphics[width=\linewidth]{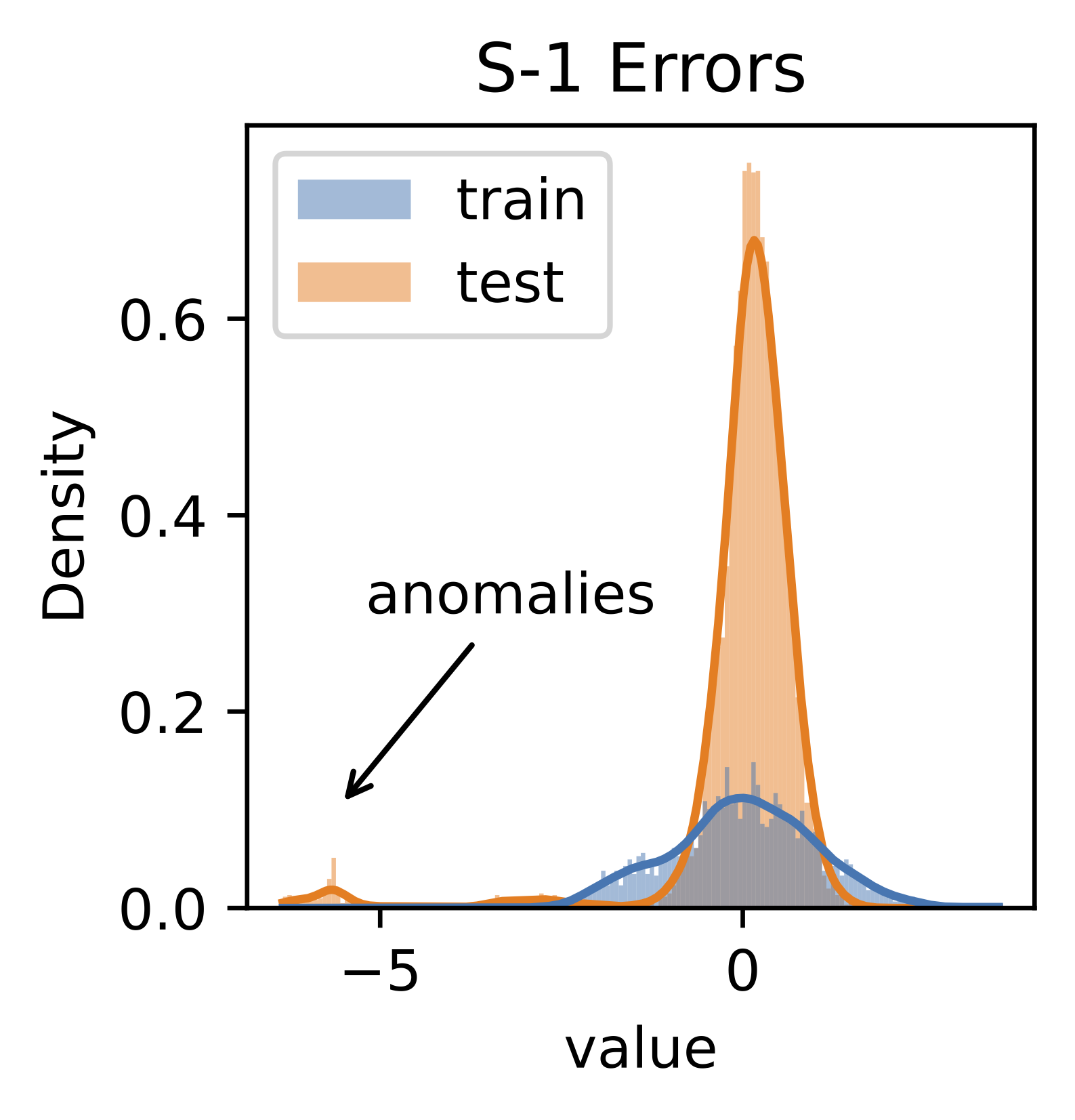}
      \caption{}
      \label{fig:s-1-hist}
    \end{subfigure}%
    \quad
    \begin{subfigure}{.445\linewidth}
      \centering
      \includegraphics[width=\linewidth]{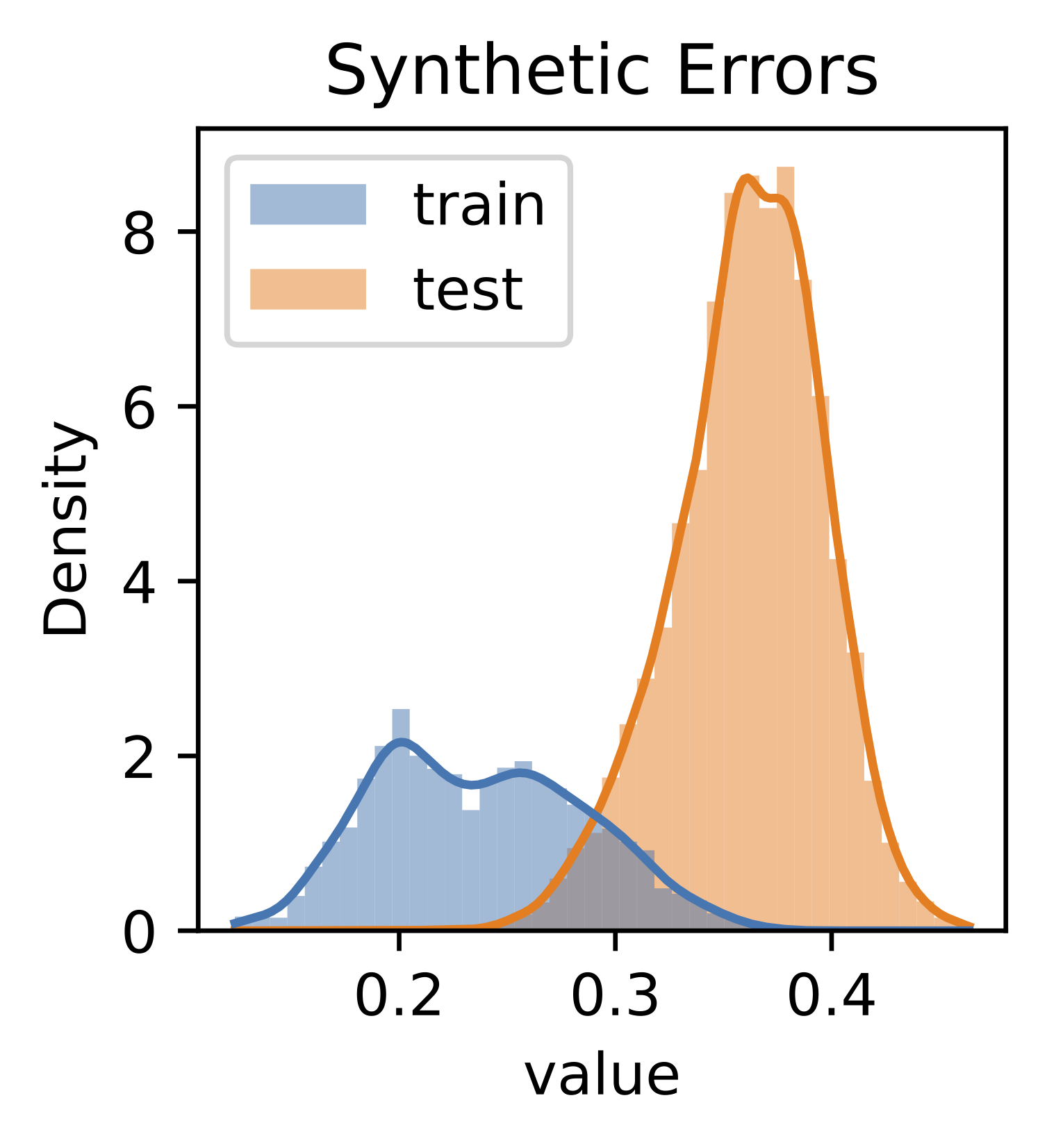}
      \caption{}
      \label{fig:random-errors}
    \end{subfigure}
    \caption{Model errors on (a) S-1 signal from SMAP dataset, where the anomalous errors are clearly separated from normal errors. (b) synthetic data were the testing error values do not have a similar distribution to training errors.}
    \label{fig:hist-errors}
\end{wrapfigure}

\section{Limitations}

While we assume that the training data is free from anomalies and both the training and testing data exhibit similar patterns, that it is not always true in practice. If the errors observed follow our assumptions, we see a case similar to Figure~\ref{fig:s-1-hist} where there are error scores that greatly deviate from the expected error observed during the training phase.
On the other hand, when the testing data differs significantly, like we observe in Figure~\ref{fig:random-errors} of a synthetically generated time series, a larger number of observations will be marked as ``anomalous''. This is a clear limitation of our model, triggering a need to retrain the origin model to conform to the new distribution of the data.



\end{document}